\documentclass[letterpaper, 10 pt, conference]{ieeeconf}  % Comment this line out if you need a4paper

\IEEEoverridecommandlockouts                              % This command is only needed if 
\usepackage{graphicx} % for pdf, bitmapped graphics files
\usepackage{subfigure}
\usepackage{url}
\usepackage{array}
\usepackage{amsmath}
\newcolumntype{I}{!{\vrule width 1pt}}
\newlength\savedwidth
\newcommand\whline{\noalign{\global\savedwidth\arrayrulewidth
		\global\arrayrulewidth 1pt}%
	\hline
	\noalign{\global\arrayrulewidth\savedwidth}}
\newlength\savewidth

\usepackage{color}

\title{\LARGE \bf
	{Realtime Global Attention Network for Semantic Segmentation}
}

\author{Xi Mo$^{1, *}$, Xiangyu Chen$^{1}$ % <-this % stops a space
	\thanks{*Corresponding Author, email: x618m566@ku.edu}% <-this % stops a space
	\thanks{$^{1}$School of Engineering, University of Kansas, Lawrence KS 66049, USA}%
}

\begin{document}

	\maketitle
	\thispagestyle{empty}
	\pagestyle{empty}
	
	\begin{abstract}
		
	 In this paper, we proposed an end-to-end realtime global attention neural network (RGANet) for the challenging task of semantic segmentation. Different from the encoding strategy deployed by self-attention paradigms, the proposed global attention module encodes global attention via depth-wise convolution and affine transformations. The integration of these global attention modules into a hierarchy architecture maintains high inferential performance. In addition, an improved evaluation metric, namely MGRID, is proposed to alleviate the negative effect of non-convex, widely scattered ground-truth areas. Results from extensive experiments on state-of-the-art architectures for semantic segmentation manifest the leading performance of proposed approaches for robotic monocular visual perception.
		
	\end{abstract}
	

\section{\MakeUppercase{Introduction}}\label{sec:intro}

Correct bin picking by suction from cluttered environment is nontrivial for a robotic hand~\cite{zeng2018robotic}. Since robotic hands don't have much prior knowledge of spatial shape by category, texture of material or normal vectors of surfaces, it remains to be an open topic whether a robotic hand is capable of recognizing and analyzing objectiveness as human being does. This topic becomes more intricate given the constraint that, the very limited information offered is the RGB images of cluttered scene, and plausible areas for suction annotated by various human experts. Individuals may make mistakes while annotating images in their unique styles, which diversifies annotations greatly, making it more difficult for traditional machine vision techniques to forge ahead with.

In terms of neural network based semantic segmentation, realtime visual perception with light computational liability is preferred, especially for robotic bin picking by suction. Zeng~et~al~\cite{zeng2018robotic} names semantic segmentations of adsorbability as affordance maps to indicate the possibilities of objects being picked up. It is widely known that, the predicted affordance maps are not yet applicable enough for actual bin picking - post-refinement of estimating normal vectors of 3D surfaces, registering affordance maps to multiple coordinates systems, calibrations on robotic hand and cameras, and data stream synchronization etc. are necessary though time-consuming. In this case, predict more reliable affordance maps in real-time with low-cost inferential processors will greatly benefit the real world implementations.

\begin{figure}[t]
	\centering
	\includegraphics[width=\linewidth]{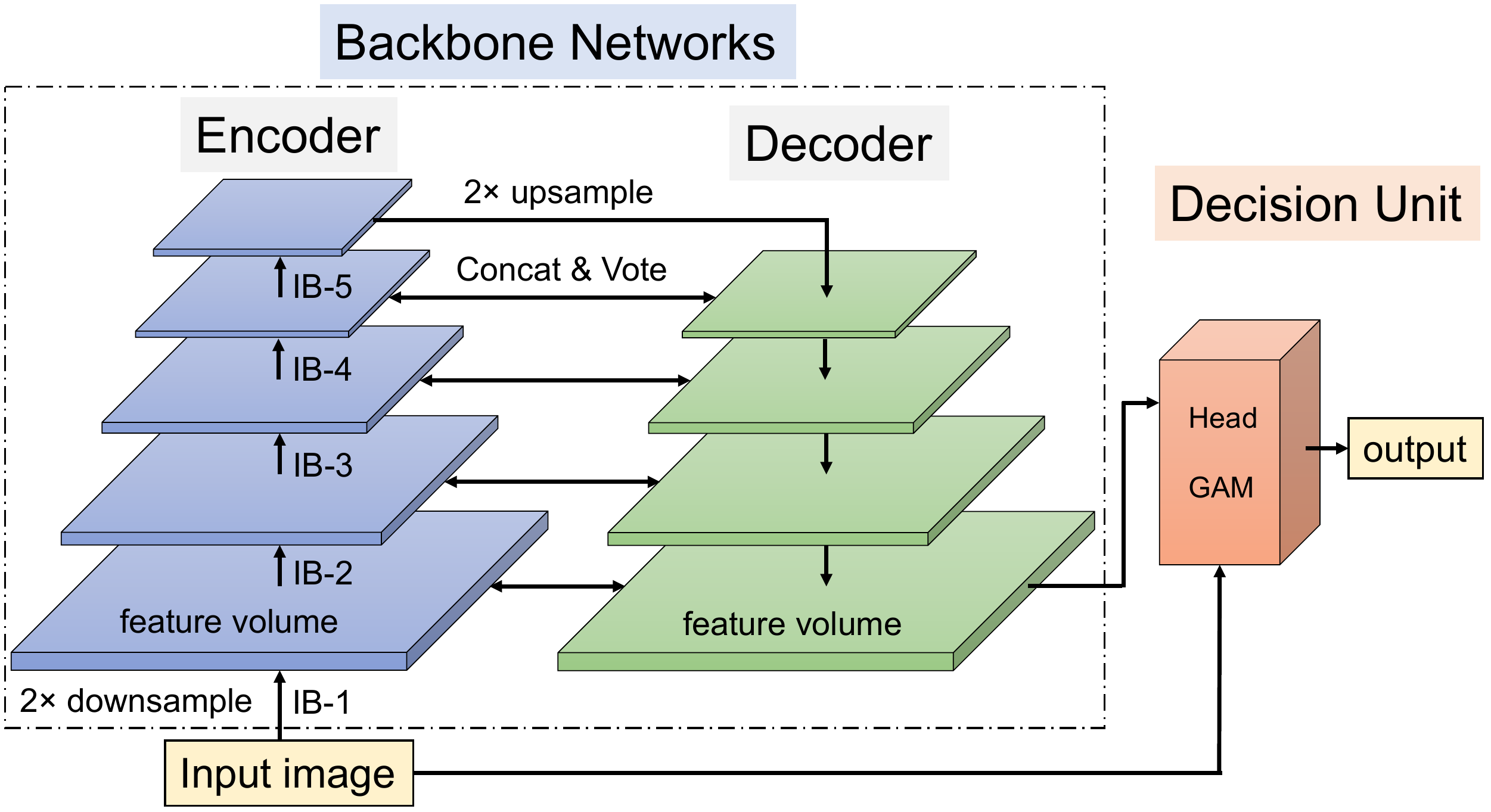}
	\caption{Hierarchy inference architecture of 5-scales RGANet5. Backbone network consists of encoder and decoder networks. One IB processes feature volume at a scale, decoder network interacts with encoder at all scales, generates high-order feature volumes. The decision unit has a Head to group both high-order features and raw image channels, and a GAM to acquire the final segmentations (affordance maps).}
	\label{fig:1}
\end{figure}

\begin{figure*}[t]
	\centering
	\includegraphics[width=\textwidth]{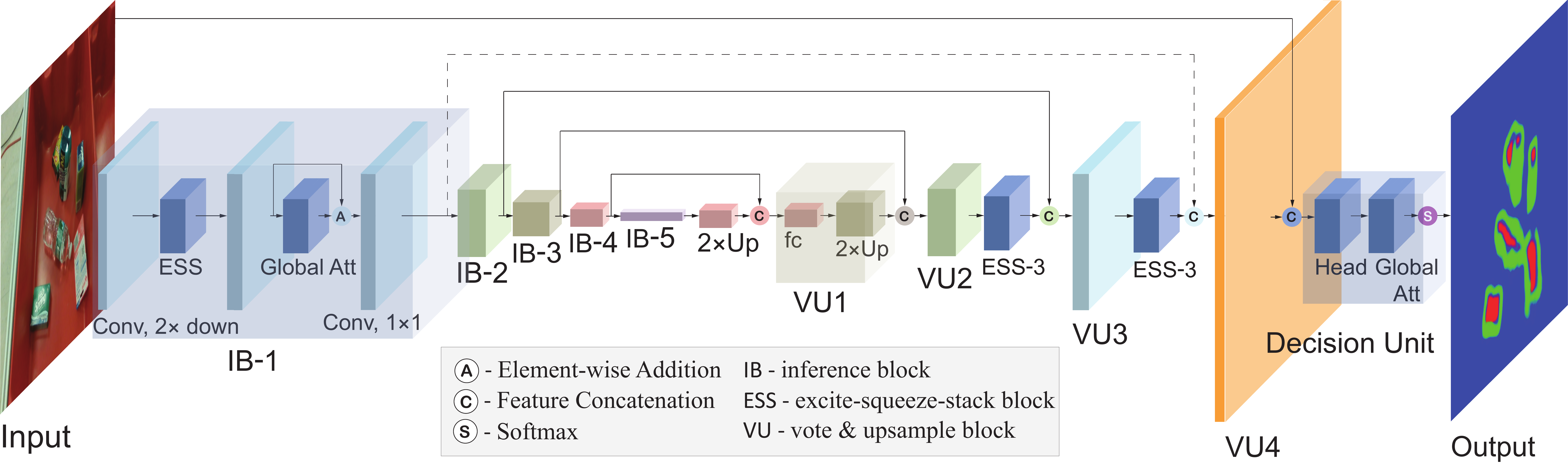}
	\caption{Forward pass of RGANet5. It should be noted that detailed framework of ESS, GAM and decision unit are presented in Section~\ref{sec:method}, thereby corresponding blocks (in deeper blue color) do not represent feature maps. The network takes monocular input image, outputs pixel-wise, one-hot encoded predictions of background (blue), negative sample (green), suction areas (red) respectively. The dashed line indicates the highway between encoder and decoder can be disconnected.}
	\label{fig:2}
\end{figure*}

We propose to efficiently predict affordance maps by realtime global attention network (RGANet), which can be easily adapted to other semantic segmentation tasks. As depicted in Fig.~\ref{fig:1}, RGANet is a light-weighted hierarchy architecture composed of inference blocks (IBs) that are based on attention mechanism~\cite{huang2019ccnet, vaswani2017attention}. For the purpose of preserving full-size activation of feature maps, pooling and dropout layers are deprecated. To explore the potentials of proposed global attention modules (GAMs), we adopt a GAM-enhanced encoder-decoder backbone network.

One specific forward pass of the proposed framework is illustrated in Fig.~\ref{fig:2}. RGANet5 has 5 levels of inference blocks (IBs), each one applies to a 2$\times$ down-sampling of its previous feature map to capture features at each scale. The decoder network is composed of vote \& upsample (VU) blocks and excite-squeeze-stack (ESS) bottleneck blocks for better decoding capabilities. Due to the enhancement of GAM at each IB, we only utilize 3$\times$3 standard convolution w/o pooling layers to downsample features, and $1\times1$ convolution for reformulating feature depth. We also tested large kernels for down-sampling, which doesn't outperform 3$\times$3 convolutions.  

Vanilla convolution convolves and aggregates feature maps faster than fully-connected layers due to shared weights. Depth-wise convolution~\cite{chollet2017xception} further reduces parameters by convolving groups of feature maps separately then concatenates. We propose to reformulate self-attention mechanism by implementing long convolutional kernels (please refer to section~\ref{sec:rga}) and depth-wise convolutions instead of computing cosine similarities of feature vectors~\cite{vaswani2017attention} or pure matrix multiplication~\cite{huang2019ccnet} upon feature volumes. Our perception is that single convolutional kernel only needs to encode neighboring features other than entire feature volume, which can be mutually correlated by affine-transforming into global encoding. This is beneficial for reducing the computational complexity, and speeding up inference to meet realtime requirements.

Furthermore, during evaluation phase, due to unconnected ground-truth areas that are distributed across the entire image per the trade-off circumstance illustrated in Fig.~\ref{fig:4b}, generally applied segmentation metrics fail to correctly evaluate predictions. Mean-Grid Fbeta-score (MGRID), a novel metric for segmentation evaluation, alleviates the flaw by two-stage operations: partition and synthetics.

To summarize our contributions, this paper highlights the following novelties:

{\noindent \bf\romannumeral1.}~Propose a one-stage hierarchy inference architecture for semantic segmentation without any auxiliary losses.

{\noindent \bf\romannumeral2.}~Propose the GAM for realtime inference - 54fps on a GTX 1070 laptop, and 134fps with a V100 GPU.

{\noindent \bf\romannumeral3.}~Propose the metric MGRID for evaluating widely-distributed predictions.

\section{\MakeUppercase{Related Works}} \label{sec: related}

\subsection{Predict Affordance Maps} \label{sec: affordance}
Zeng~et~al~\cite{zeng2018robotic} implemented two FCNs~\cite{long2015fully} with ResNet-101~\cite{szegedy2017inception} backbones to fuse color and depth streams of cluttered scene. This approach yields 83.4\% precision at Top-1\% percentile affordance proposals. Azpiri~et~at~\cite{azpiri2021affordance} further refined the Top-1\% percentile precision to $\sim$94\% by a deep Graph-Convolutional Network backbone, which outperforms FCN (ResNet-101 backbone) by $\sim$2\%. Neither of two approaches takes colored images as the only input, consuming much inference time in processing depth information. Shao~et~at~\cite{shao2019suction} proposed to predict affordance map by sharing a ResNet-50 backbone between colored image stream and depth image stream, and a U-Net~\cite{ronneberger2015u} to fuse features at different scales. Their approach is annotation-free, but requires the robotic hand attempting to find the most possible points for suction, which is more expensive than deriving direct predictions from monocular images.

\subsection{Self-Attention Modules in Semantic Segmentation} \label{sec: attention}
Self-Attention mechanism was first proposed in the field of nature language processing~\cite{vaswani2017attention} for temporal domain. Wang~et~al~\cite{wang2018non} adapted the idea to spatial domain, that a feature vector is spatially related to all other features in the feature volume, the features that are highly related will generate strong response, which facilitates network to model the non-local relations across the entire volume.

Li~et~al~\cite{li2018pyramid} designed a fully-convolutional feature pyramid attention module to replace the spatial-pyramid-pooling module~\cite{zhao2017pyramid}, and a global-attention-upsample module with which high-level features perform global average pooling~\cite{zhou2016learning} as the guidance for low-level features. Woo~et~al~\cite{woo2018cbam} believed that simple channels-wise attention and spatial-wise attention sub-modules can boost representation power of CNNs. Recently, OCNet~\cite{yuan2021ocnet} efficiently aligns a global relation module and a local relation module, dividing and merging feature volumes in different styles. Bello~et~al~\cite{bello2019attention} proposed to augment standard convolution by attention mechanism. Cao~et~at~\cite{cao2019gcnet} proposed a global-context block that benefits from simplified non-local block~\cite{wang2018non} and squeeze-excitation module~\cite{hu2018squeeze}. The proposed GAM in this paper is inspired by the criss-cross attention (CCA) mechanism of CCNet~\cite{huang2019ccnet}. CCNet approximates non-local attention by two cascade CCA modules, each of which correlates all feature vectors aligned horizontally and vertically. 

\begin{figure*}
	\centering
	\subfigure[Illustration of ESS-3 framework (3 BottleNecks).]{
		\label{fig:3a}
		\includegraphics[width=0.4\textwidth]{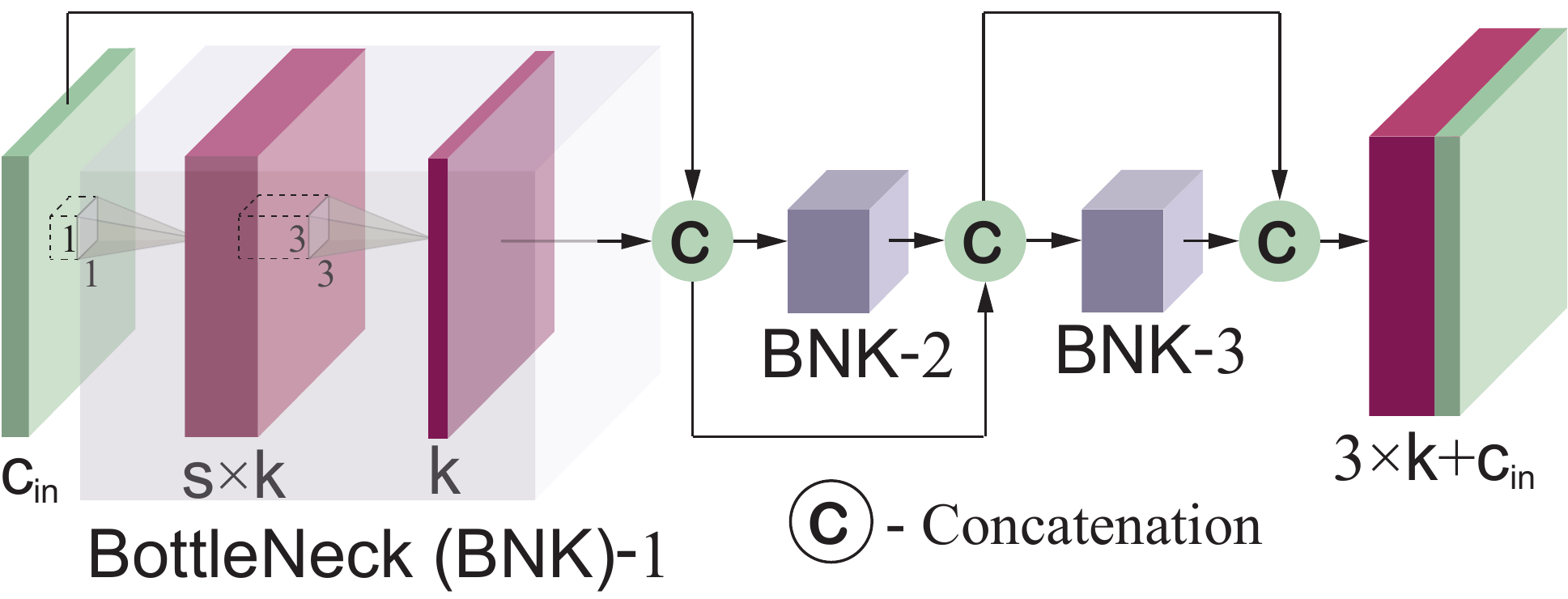}}
	\subfigure[The structure of GAM.]{
		\label{fig:3b}
		\includegraphics[width=0.58\textwidth]{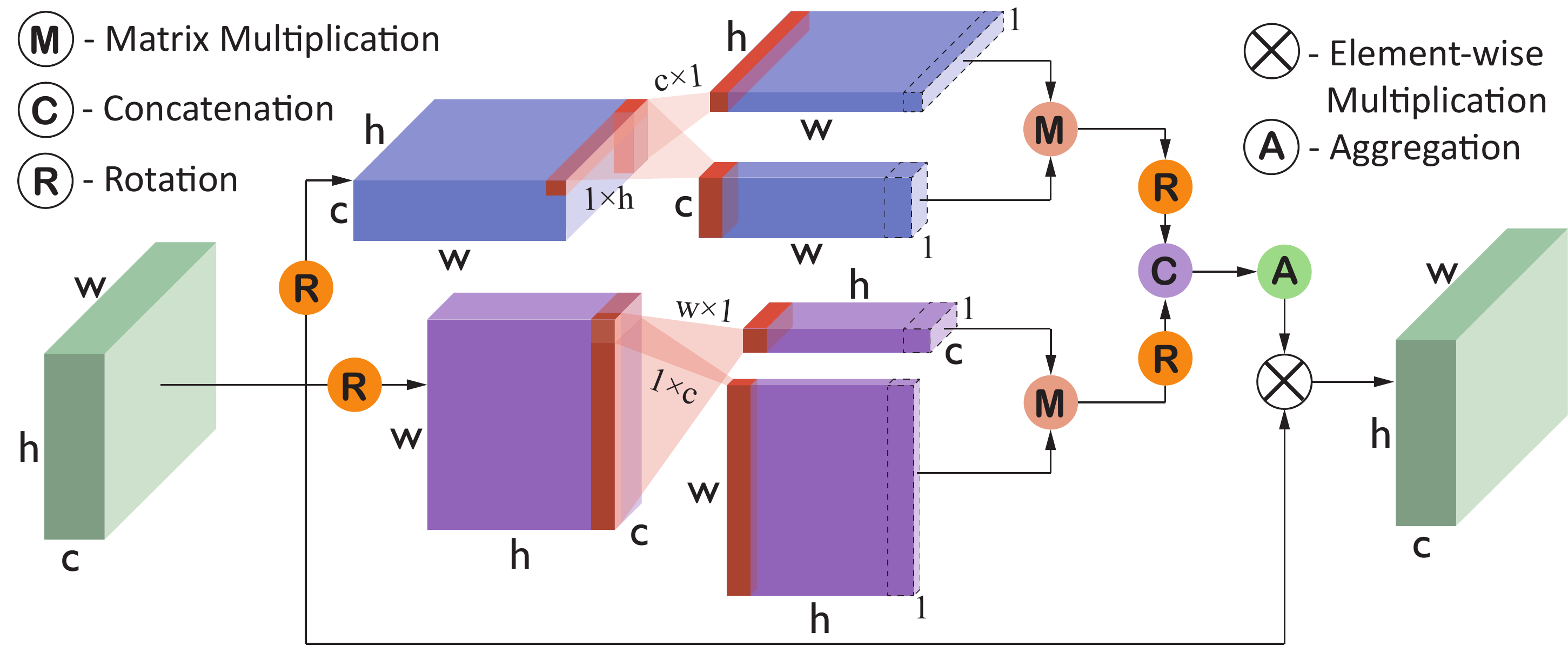}}
	
	\caption{ESS-3 and realtime GAM. (a) ESS-$n$ block first expands input channel $C_{in}$ channels to $s\times k$ channels by 1$\times$1 convolution, scalers \emph{s} and \emph{k} control the number of expanded channels; then it squeezes intermediate features to $k$ channels by $3\times 3$ convolution, densely stacking all $k$-channel features gradually towards final $n\times k + C_{in}$ channels. Specially, we set $C_{in}=k$ to enforce the entire network scalable w.r.t. ratio $k$. (b) Query and key of GAm are derived from affine transforms, either query or key is channel-wise ($1\times c$ and $c\times 1$ depth-wise convolutions) and spatial-wise ($w\times 1$ and $1\times h$ depth-wise convolutions), globally related by matrix multiplication and aggregation layers.}
	\label{fig:3}
\end{figure*}

\section{\MakeUppercase{Methodology}} \label{sec:method}

\subsection{Hierarchy Inference Architecture} \label{sec:arch}
RGANet5 is designed as a single `distillation tower' (see Fig.~\ref{fig:2}) with 5 temperature levels, each level is composed of one IB. The architecture is expandable such that RGANet$n$ has $n$ cascade IBs. Each IB halves input feature size, then modulates channels to ratio $k=15$ by a standard $3\times3$ convolution. ESS block (see Fig.~\ref{fig:3a}) is regarded as the backbone for IB, we adopt ESS-3, ESS-6, ESS-12, ESS-24 for IB-1 and IB-2, IB-3, IB-4, IB-5 respectively. Output of GAM (refer to Fig.~\ref{fig:3b}) is directly added to the output feature volume $\textbf x$ of ESS block as the residual $\lambda \cdot \textbf x$, which formulates the incremental up-sampling layers accordingly. The final output takes the form $(1+\lambda) \cdot \textbf x$, where $\lambda$ is the learnable weights volume that has the same size of $\textbf x$, and operator `$\cdot$' signifies element-wise product. Therefore, non-negative activation of GAM artifact is preferred, and Batch-Norm (BN) layers are necessary to restrict its upper bound.

As the synthesis of 5-scales distillation artifacts, highway connections not only populate previous inferences to up-sampling layers accordingly, they also facilitate gradient back-propagation during training phase, especially for a very deep network. The `vote \& upsample' (VU) block (see Fig.~\ref{fig:2}) weights concatenated input feature maps by $1\times 1$ convolution without bias, these weighted features are then $2\times$ upsampled via nearest interpolation. If inferring without last two ESS-3 blocks, RGANet will not yield fine-grained predications. Also, we noticed that IB-1 is directly linked to the last UV4 block, which performs poorly without Decision Unit (DU) shown in Fig.~\ref{fig:2}. DU consists of feature modulating head to deduct channels from $k+3$ to the number of classes by $1\times 1$ convolution, and one last GAM that yields predictions. 

\begin{figure*}[t]
	\centering
	\subfigure[]{
		\label{fig:4a}
		\includegraphics[height=95px]{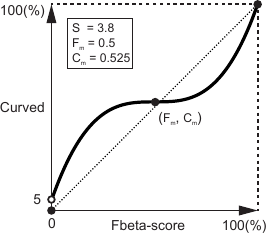}}
	\subfigure[]{
		\label{fig:4b}
		\includegraphics[width=0.76\textwidth]{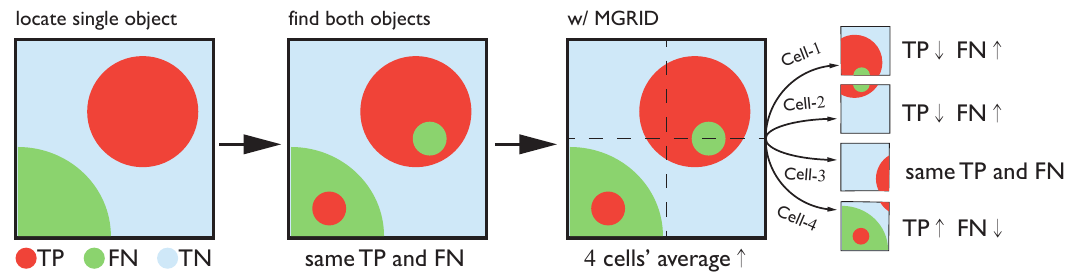}}
	
	\caption{MGRID metric. (a) Synthetic function. (b) A special case for predicting suction areas when the same precision and recall indicate better predictions for real-world implementation, and how MGRID mitigates the circumstance by partition and synthetics. True Negative (TN) and False Positive (FP) samples stay unchanged during the processes.}
	\label{fig:4}
\end{figure*}

\subsection{Realtime Global Attention} \label{sec:rga}

Fig.~\ref{fig:3b} illustrates the pipeline of the GAM. With a batch-size of 1, input feature volume $\bf{x}$ filtered by previous ESS block has a shape of $h\times w \times c$, which is rotated to $c\times h \times w$ query $Q(c, h, w)$ (upper branch, blue color) and $w\times c \times h$ key $K(w, c, h)$ (lower branch, purple color). Query conducts channel-wise attention via depth-wise sliding $w$ kernels $W(c, 1)$ shaped $c\times 1$ across $c-h$ view, which results in a $c\times 1 \times w$ feature volume $Q^{c}$. Similarly, $h$ horizontal positions are encoded by $1\times h$ depth-wise convolutional kernel $W(1, h)$, mapping into the $c\times 1 \times w$ horizontally positional encoding $Q^{h}$. Key is transformed accordingly except the sizes of kernels, we denote its artifacts as channel-wise encoding $K^{c}$ and vertically positional encoding $K^{v}$. The final output $F_{out}$ are then formulated as:
\begin{equation}\label{eqn:1}
	F_{out} = \lambda \cdot \textbf{x} = f\{\textrm{rot}[Q^{h}Q^{c}]~\bigcup~\textrm{rot}[K^{v}K^{c}]\}\cdot\textbf{x},
\end{equation}
where $\textrm{rot}[\cdot]$ operator signifies rotation; $f\{\cdot\}$ operator performs $1\times 1$ convolution to resize $2c$-channels to $c$-channels with Swish activation function~\cite{Ramachandran2018searching}, then maps to [0, 1] weights volume via Sigmoid or Softmax functions; `$lhs~\bigcup~rhs$' operator concatenate $lhs$ and $rhs$ features; Let `$*$' denotes the depth-wise convolution, we know:
\begin{equation}\label{eqn:2}
	\begin{aligned}
		Q^{h} = Q(c, h, w) * W(1, h), & Q^{c} = Q(c, h, w) * W(c, 1), \\
		K^{v} = Q(w, c, h) * W(w, 1), &K^{c} = Q(w, c, h) * W(1, c).
	\end{aligned}
\end{equation}

Consider Eqn.~\eqref{eqn:1} and Eqn.~\eqref{eqn:2}, RGA's total number of trainable parameters is $2hw+cw+hc$ (exclude BN layer, bias and $1\times 1$ point-wise convolution), while for vanilla convolution with the same kernel height and width, this number becomes $hw^2+wh^2+cw^2+ch^2$. In terms of global attention, although matrix multiplication operation only correlates features horizontally and vertically aligned for any feature vector, $Q^{h}$, $Q^{c}$, $K^{v}$ and $K^{c}$ themselves are the artifacts of global depth-wise convolutions, which, as a result, each element in these 4 Queries and Keys bounds other elements with shared weights. We can also treat RGA as the type of `learnable global attention'. Furthermore, all affine operations of RGA module are differentiable, we didn't observe vanishing, or exploding gradients issues during training.  

\subsection{Rethink Densely-Stacked Bottlenecks} \label{sec:ess}

A Dense block~\cite{huang2017densely} has multiple densely-connected Bottlenecks (BNK) modules, each BNK acts in a stack-squeeze manner, and the last one outputs a $k$-channels feature volume.

The ESS-$n$ block, as illustrated in Fig.~\ref{fig:3a}, outputs a stacked $(n+1)k$-channels feature volume by $n$ densely-connected BNKs, each contributes $k$-channels of features, and we let $C_{in}=k$.  Each channel of stacked feature maps shows one degree of filtered raw input features, such that RGA modules are capable of correlating high-order features to low-level features. RGANet5 adopts ESS-3 for IB-1, ESS-3 for IB-2, ESS-6 for IB-3, ESS-12 for IB-4, and ESS-24 for IB-5. One of our future works is to let RGA module ignore noisy low-level features, and locate more reliable high-order features.

\subsection{MGRID metric for Evaluation} \label{sec:mgrid}

Existing evaluation metrics poorly treat the special case of a prediction map as illustrated in Fig.~\ref{fig:4b} that, when original predictions cover object-1 (red), but fail to locate object-2 (green) at the left-bottom corner of image, off-the-shelf metrics yields the same results if part of object-1 relocates to the object-2, because this relocation does not affect paranormal statistics of TP, FN, TN and FP. In reality, we want predications to be able to cover more objects, such that a robotic hand would seek-and-pick all objects even though Precision or Recall is yet not favorable enough (e.g., $\sim$15\% Recall on object-2 alone). It would be more reasonable to assign higher score for the prediction map that covers 2 objects. 

\paragraph{Partition.}2-Stages MGRID metric aims to remedy the issue mentioned above. As shown in the third image of Fig.~\ref{fig:4b}, during partition stage, predication map is manually divided into four cells, each cell is treated fairly against any other cell. An ideal partition will separates objects by different cells. Next, only calculate Fbeta-scores (or any other existing metrics) for all cells that contain predictions and categorical ground-truth (non-zero TP, FP or FN samples), using the definition
\begin{equation}
	F_{\beta} = \frac{(1+\beta^2)TP}{\beta^2(TP+FN)+TP+FP+\epsilon},
\end{equation}
where $\epsilon=1\times10^{-31}$, $\beta >1$ weights Recall more than Precision, and $ 0<\beta<1$ weights the opposite. 

\paragraph{Synthetics.} The second step is to synthesize all $n$ Fbeta-scores $\mathcal{F}=\{F_{\beta}(i)|i=1,2,...,n\}$ collected from all partitions, any Fbeta-score $F\in \mathcal{F}$ is curved by a regulator as shown in Fig.~\ref{fig:4a}, which takes the form
\begin{equation}
	\Gamma (F) = 
	\left\{
	\begin{array}{ll}
		S(F-F_{m})^3+C_m, & 0<F\leq 100\%\\
		0, & F=0
	\end{array}\right.,
\end{equation}
where coefficients $0<F_m,~C_m<1$ requires to be manually set. Fig.~\ref{fig:4a} shows the curve by setting $(F_m, C_m) =(0.5, 0.525)$, then $S$ is calculated as $S=(1-C_m)/(1-F_m)^3$. Let $T = [F_m/(1-F_m)]^3$, then intercept $B$ at $F=0$ can be denoted as $C_m(1+T)-T$. To make the regulator effective, $B$ should fall within the interval $(0, F_m)$, which leads to valid ranges of $F_m$ and $C_m$:
\begin{equation}
	\frac{T}{1+T}<C_m <\frac{F_m+T}{1+T},~0<F_m<1.
\end{equation}
The final confidence score is derived by
\begin{equation}
	MGRID = \frac{1}{n}\sum_{F\in\mathcal{F}} \Gamma(F).
\end{equation}

\section{\MakeUppercase{Experiments}} \label{exp}

\subsection{Configuration} \label{sec:dataset}
Public-available suction dataset~\cite{suctiondataset} consists of camera intrinsics and pose records, RGB-D images of clutter scenes and their backgrounds, and labels. We only adopted color images and labels w/ a train-split of 1470 images and a test-split of 367 images, each image has a resolution of 480$\times$640. All colored images were normalized to tensors valued between 0 and 1, which were not resized or padded during training and testing.

\begin{figure*}[t]
	\centering
	\includegraphics[width=0.8\textwidth]{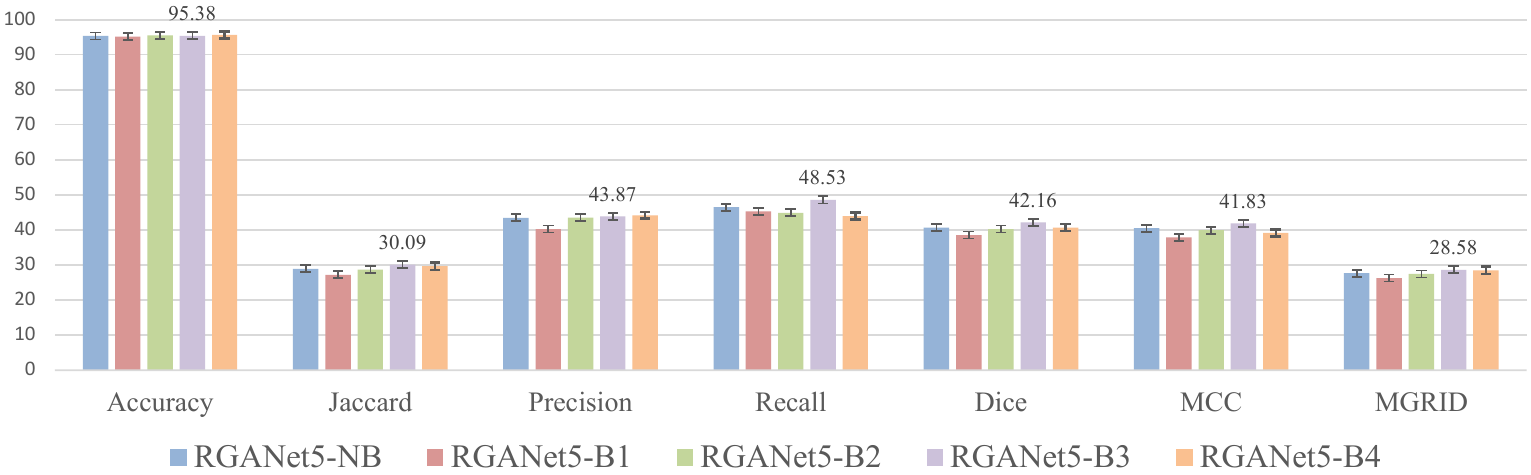}
	\caption{Averages and its estimated standard deviations over test-set w/o DU. All values are presented in `\%'. RGANet5-B$n$ signifies last $n$ highways are blocked. RGANet5-B4 (purple) shows the best performance among all tested architectures. }
	\label{fig:5}
\end{figure*}

\subsection{Train and Test} \label{sec:train_test}

We implemented AdamW optimizer~\cite{loshchilov2018decoupled} with AMSGrad~\cite{reddi2018on}, compared two weighted loss function during training - focal loss (FLoss)~\cite{lin2017focal} and cross-entropy loss (CELoss). Assume $y$ denotes the prediction, $y'$ the one-hot encoded ground-truth, $n$ the total number of classes, $\gamma$ and $\alpha$ are constants, then FLoss and CEloss are denoted as:

\begin{equation}
	\begin{aligned}
		&FLoss(y, y') = -\sum_{c=1}^{n} \alpha_{c}\sum_{c} (1-y)^{\gamma}\log y,\\ &CELoss(y, y') = -\sum_{c=1}^{n} \alpha_{c}\sum_{c} \log y
	\end{aligned}
\end{equation}

where $\sum_{c=1}^n \alpha_c=1$ and $\alpha_c \geq 0$, $\gamma > 0$, and all $y\in[0, 1]$. Training-set is augmented during training by random hue, flip, rotation, blur, shift etc.

Online testing comes simultaneously during training, which shows calculations of Jaccard Index, precision, and recall of running batches. The Offline testing loads checkpoint, merely evaluates the class corresponds to predicted suction areas.

All experiments were conducted using a GTX1070 laptop, and one Tesla V100 GPU. Network scaler $k$ was set to 15 constantly for a better trade-off between module size and performance. We set constant learning rate to $1.5\times 10^{-4}$, weights decay rate to 0; $\gamma=1.3$, $\alpha_1=0.25$ and $\alpha_2=0.25$ upon background and negative samples, $\alpha_3=0.5$ upon suction areas because of unbalanced proportions; default MGRID parameters $\beta=0.5$, grid intervals $(\delta_H, \delta_W) = (12px, 12px)$, $C_m$ and $F_m$ took the same values as illustrated in Fig.~\ref{fig:4a}. The training of RGANet5 lasted for $\sim$2 days.

We implement multiple metrics to evaluate inferential artifacts. Note that for the suction dataset, users only care about feasible regions to adhere. Therefore, only the category that represents predicted suction areas is evaluated, the final scores are computed via averages over entire test-set.

\subsection{Ablation Study} \label{sec:block}

\paragraph{Block highways between IBs and remove DU}To evaluate the performance of RGANet5 w/o DU, as well as the effects of disconnecting different highways between IBs, we tested 5 circumstances when different highways to IBs are blocked. The results were acquired via CELoss because of its better performance than implementing FLoss. We find that, FLoss enhances weights for negative samples to weaken salient positive responses, in the meantime, it also `confuses' inferential and voting blocks of RGANet by unreasonable probabilistic distributions of classes, especially when negative samples dominate. Additionally, we applied nearest interpolation instead of deconvolutional layers to VU.

Fig.~\ref{fig:5} illustrates the testing performance of RGANet5-NB, RGANet5-B1, RGANet5-B2, RGANet5-B3, and RGANet5-B4 by comparing mean and observing standard deviation of accuracy, Jaccard Index, precision, recall, Dice (F1-score), MCC (Phi Index) and MGRID. It is obvious that IB-1 provides the most crucial lower level features, IB-2, IB-3, IB-4 and IB-5 refine those features distinctively - IB-4 and IB-5 provide more useful semantic information than IB-2 and IB-3. When all highways are disconnected, RGANet5-B4 outputs less precise proposals than RGANet5-B3 - the best performer during the test.

In conclusion, blocking highways has insignificant influence on the performance over test-set.

\begin{table}
	\renewcommand{\arraystretch}{1.4} 
	\begin{center}
		\setlength{\tabcolsep}{2,4mm}{ 
		\begin{tabular}{l|c|c|c|c|c}
			\whline
			Method & Prec. & Recall & Jacc. & Dice & MGRID \\\hline
			RGANet-B3(w/o) & 43.87 & 48.53 & 30.09 & 42.16 & 28.58 \\
			RGANet-B3(w/) & \bf{50.52} & 58.92 & 37.84 & 50.44 & 34.20 \\ 
			RGANet-NB(w/o) & 43.45 & 46.45 & 28.88 & 40.65 & 27.65 \\
			RGANet-NB(w/) & 47.16 & \bf{63.36} & \bf{37.92} & \bf{50.60} & \bf{34.58} \\
			\whline
		\end{tabular}}
	\end{center}
	\caption\small{Ablation study on DU. All averaged evaluation metrics are presented in `\%'. GAM in DU tends to make better trade-off between precision and recall.}
	\label{tab:1}
\end{table}

\paragraph{Without DU vs With DU}

According to Tab.~\ref{tab:1}, GAM boosts the overall performance of the backbone networks by large margins, indicating the powerful encoding capabilities of global attention mechanism. Furthermore, GAM shows a favor of correlating more feature maps, which leads to the superior performance of RGANet-NB. As for RGANet-B3(w/ DU), we also substituted the nearest interpolation layer in VU by a $2\times 2$ deconvolutional layer to evaluate its up-sampling performance. 

\begin{figure*}[htbp]
	\centering
	\includegraphics[width=\textwidth]{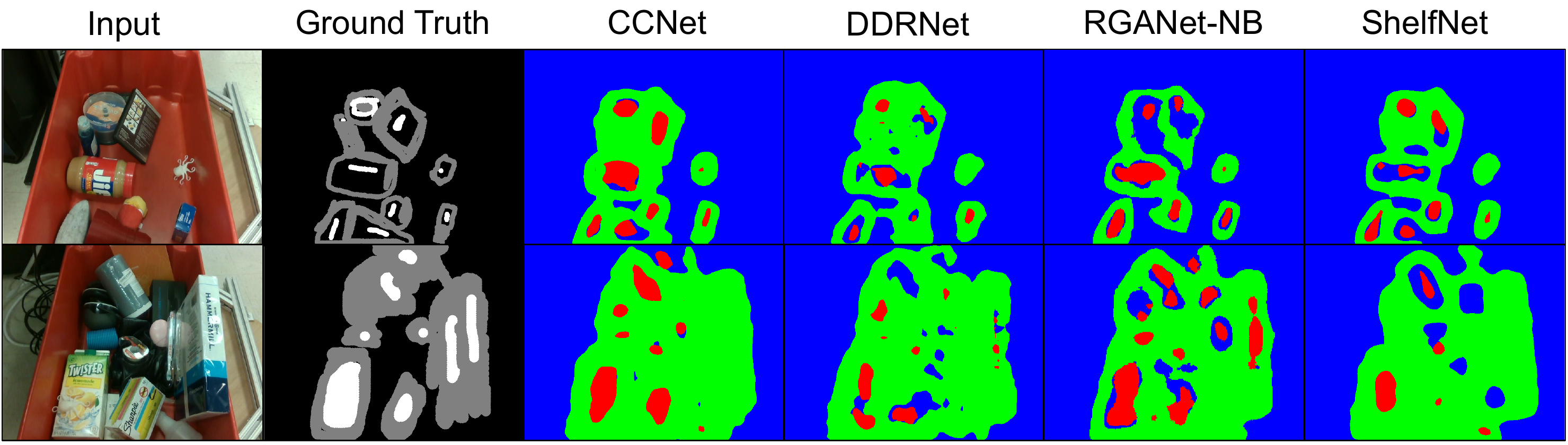}
	\caption{Qualitative results on test-set. Red regions indicate the highest affordance for adhesion, which corresponds to white regions in ground-truth. }
	\label{fig:6}
\end{figure*}

\begin{table*}
	\renewcommand{\arraystretch}{1.4} 
	\begin{center}
		\setlength{\tabcolsep}{2,4mm}{ 
			\begin{tabular}{l|c|c|c|c|c|c|c}
				\whline
				Group 1 - Large Models									& Backbone 			& Parameters 	& Inference Time 		& FLOPs 	& Jaccard 	& Dice 		& MGRID \\\hline
				RGANet5-B3 (\bf Ours) 									& - 				& 3.67M 		& \bf7.45ms / 134fps 	& 1.83B 	& 37.84 	& 50.44 	& 34.20\\
				RGANet5-NB (\bf Ours) 									& - 				& \bf3.41M 		& 7.51ms / 133fps		& \bf1.57B 	& 37.92 	& 50.60 	& 34.58\\ 
				CCNet (ResNet101)~\cite{huang2019ccnet} 				& ResNet101 		& 71.27M 		& 51.03ms / 19fps 		& 73.03B 	& \bf43.83 	& \bf 57.00 & \bf 38.90  \\
				FCN (ResNet50)~\cite{long2015fully}		    			& ResNet50 			& 32.96M 		& 22.85ms /	43fps		& 32.48B 	& 42.15 	& 55.32		& 37.91  \\
				FCN (ResNet101)~\cite{long2015fully} 					& ResNet101 		& 51.95M 		& 39.41ms / 25fps		& 50.72B 	& 42.28 	& 55.51 	& 37.69 \\ 
				DeepLabv3 (ResNet50)~\cite{florian2017rethinking}		& ResNet50  		& 39.64M 		& 32.71ms / 30fps		& 38.40B 	& 43.17 	& 56.32 	& 38.42 \\
				DeepLabv3 (ResNet101)~\cite{florian2017rethinking} 		& ResNet101 		& 58.63M 		& 49.17ms / 20fps		& 56.63B 	& 41.98 	& 55.05 	& 37.55 \\
				BiSeNetv1~\cite{yu2018bisenet}							& ResNet18			& 23.08M 		& 9.65ms  / 103fp/s		& 9.53B 	& 37.64 	& 50.38 	& 34.18 \\
				\whline
		\end{tabular}}
	\end{center}
	\caption\small{Comparison with large semantic segmentation models using a Tesla V100 GPU. All averaged evaluation metrics are presented in `\%'. Proposed approaches achieve competitive performance with the least total parameters, indicating a better trade-off between model size and performance.}
	\label{tab:2}
\end{table*}

\begin{table*}
	\renewcommand{\arraystretch}{1.4} 
	\begin{center}
		\setlength{\tabcolsep}{2,4mm}{ 
			\begin{tabular}{l|c|c|c|c|c|c|c}
				\whline
				Group 2 - Light-Weighted Models						& Backbone 			& Parameters 	& Inference Time 		& FLOPs 	& Jaccard 	& Dice 		& MGRID \\\hline
				RGANet5-B3 (\bf Ours) 						& - 				& 3.67M 		& 7.45ms / 134fps 		& 1.83B 	& 37.84 	& 50.44 	& 34.20\\
				RGANet5-NB (\bf Ours) 						& - 				& 3.41M 		& 7.51ms / 133fps		& 1.57B 	& \bf37.92 	& \bf50.60 	& \bf34.58\\ 
				DeepLabv3~\cite{sandler2018mobilenetv2} 	& MobileNetv2 		& 4.12M 		& 2.92ms  / 342fps		& 1.16B 	& 34.61 	& 47.33 	& 31.76\\
				DDRNet-23-slim~\cite{hong2021deep}		 	& - 				& 5.69M 		& \bf1.23ms / 813fps 	& 1.07B 	& 32.30 	& 43.95 	& 32.30\\
				HRNet-small-v1~\cite{wang2020deep}			& -		 			& \bf1.54M 		& 1.84ms /	543fps		& \bf0.97B 	& 34.31 	& 46.28		& 31.97\\
				HarDNet~\cite{chao2019hardnet} 				& - 				& 4.12M 		& 1.63ms / 613fps		& 1.03B 	& 35.15 	& 47.13 	& 32.61\\ 
				ShelfNet~\cite{zhuang2019shelfnet}			& ResNet18 			& 14.57M 		& 2.06ms / 485fps		& 2.91B 	& 36.17 	& 48.61 	& 33.32\\
				STDCv1~\cite{fan2021rethinking}				& - 				& 14.23M 		& 2.45ms / 408fps		& 5.48B 	& 36.10 	& 48.34 	& 33.19\\
				\whline
		\end{tabular}}
	\end{center}
	\caption\small{Comparison with light-weighted semantic segmentation models using a Tesla V100 GPU. All averaged evaluation metrics are presented in `\%'.}
	\label{tab:3}
\end{table*}

\subsection{Compare to State-of-the-Arts}

We conducted experiments to compare RGANet5 with several novel semantic segmentation approaches. These approaches can be divided into two groups - one group that has much more parameters/FLOPs that substantially can outperform RGANet, and the other one that has comparable model sizes. As shown in Tab.~\ref{tab:2}, Deeplabv3~\cite{florian2017rethinking} and FCN~\cite{long2015fully} do not rely on attention mechanism, while CCNet~\cite{huang2019ccnet} has merely two cascade CCA modules that bring in tremendous amount of trainable parameters and operations. Light-weighted RGANet with 6 GAMs, on the other hand, achieves competitive performance (4-7\% less) against the best performer on the test-set. Also, RGANet5 outperforms BiSeNetv1~\cite{yu2018bisenet}, another attention-based realtime approach, by $\sim6\times$ less parameters and FLOPs.

As illustrated in Tab.~\ref{tab:3}, proposed approach achieves the best Jaccard Index, Dice coefficient and MGRID score when compared with top-tier realtime approaches selected from Cityscape Leader board~\cite{cityscape}. Although behaves better in metrics evaluation, RGANet runs relatively slow due to the fact that PyTorch is well-optimized for convolutional neural networks. It is one of our future works to further optimize RGANet for faster and more accurate inference. Readers may refer to Fig.~\ref{fig:6} for our qualitative evaluation.

\section{Conclusion and Future Works}
Firstly, we introduced a novel light-weighted, hierarchy inference network embedded with realtime global attention modules. Densely-connected excite-squeeze-stack blocks generate feature volume as the input to realtime global modules, and the attention module correlates features via learnable weights and affine transformations. Ablation study, as well as the comparison with the state-of-the-art approaches manifests the competitive performance of the proposed RGANet5. Secondly, we designed the MGRID metric, which effectively leverages on the weights of predictive regions via partition and synthesis stages. Our future works include but not limit to, enhance the encoding capability of inferential blocks by efficient backbone networks and optimizations.

% This command serves to balance the column lengths on the last page of the document manually. It shortens the textheight of the last page by a suitable amount. This command does not take effect until the next page so it should come on the page before the last. Makesure that you do not shorten the textheight too much.

\section*{ACKNOWLEDGMENT}

We would like to thank Li, Rui for the sponsorship of NVIDIA GPUs and INTEL CPUs. We appreciate authors of references~\cite{huang2019ccnet, long2015fully, florian2017rethinking, yu2018bisenet, hong2021deep, chao2019hardnet, zhuang2019shelfnet, fan2021rethinking} for providing open-source codes of their great works.
\addtolength{\textheight}{-4cm}  
\bibliographystyle{IEEEtran} % use IEEEtran.bst style
\bibliography{egbib}

\end{document}